\documentclass[10pt,twocolumn,letterpaper]{article}
\usepackage{ijcb}
\usepackage[accsupp]{axessibility}
\usepackage[export]{adjustbox}
\usepackage{times}
\usepackage{epsfig}
\usepackage{bbm}
\usepackage{graphicx}
\usepackage{amsmath}
\usepackage{amssymb}
\usepackage{import}
\usepackage{hyperref}
\usepackage{caption}
\usepackage{subcaption}
\usepackage{balance}

\usepackage{xcolor,colortbl}
\definecolor{boxgrey}{HTML}{F3F3F3}
\newcolumntype{a}{>{\columncolor{boxgrey}}r}

\newcommand{\hlbox}[2]{
  \begin{center}
    \fcolorbox{white}{boxgrey}{
      \parbox{0.9\columnwidth}{\noindent \textbf{#1}. \textit{#2}}
    }
  \end{center}
}

\ijcbfinalcopy 

\ifijcbfinal\fi
\begin{document}

\title{(Un)fair Exposure in Deep Face Rankings at a Distance}

\author{Andrea Atzori\\
University of Cagliari\\
V. Ospedale 72, Cagliari, Italy\\
{\tt\small andrea.atzori@unica.it}
\and
Gianni Fenu\\
University of Cagliari\\
V. Ospedale 72, Cagliari, Italy\\
{\tt\small fenu@unica.it}
\and
Mirko Marras\\
University of Cagliari\\
V. Ospedale 72, Cagliari, Italy\\
{\tt\small mirko.marras@acm.org}
}

\maketitle
\thispagestyle{empty}

\begin{abstract}
  Law enforcement regularly faces the challenge of ranking suspects from their facial images. Deep face models aid this process but frequently introduce biases that disproportionately affect certain demographic segments. While bias investigation is common in domains like job candidate ranking, the field of forensic face rankings remains underexplored. In this paper, we propose a novel experimental framework, encompassing six state-of-the-art face encoders and two public data sets, designed to scrutinize the extent to which demographic groups suffer from biases in exposure in the context of forensic face rankings. Through comprehensive experiments that cover both re-identification and identification tasks, we show that exposure biases within this domain are far from being countered, demanding attention towards establishing ad-hoc policies and corrective measures. The source code is available at \url{https://github.com/atzoriandrea/ijcb2023-unfair-face-rankings}. 

\end{abstract}

\section{Introduction}
The ubiquity of rankings as a communication tool across digital platforms is, beyond doubt, offering a salient method for conveying information to users. This prevalence spans a spectrum of domains, ranging from search engines \cite{gao2020toward} and online marketplaces \cite{wan2020addressing} to music platforms \cite{pereira2019online} and news feeds \cite{sanchez2020easing}. Another notable context that heavily relies on rankings is information forensics, particularly within systems based on facial biometrics \cite{jacquet2020automated}. Operating within the spheres of law enforcement, government, and surveillance, ranking systems are extensively employed for tasks encompassing the identification of suspects in lineups and the intricate endeavor of identifying or re-identifying individuals through video surveillance data. This prominence of rankings is increasingly underpinned by the integration of deep face recognition models, distinguished for their impressive accuracy. These models are improving system capabilities, by potentially reducing errors in suspect identification.

From a biometric perspective, the literature extensively illustrates how deep learning models, along with the latent representations they yield, lead to higher failure rates for certain demographic groups~\cite{mitigate_racial,genderineq,sexist,DBLP:conf/interspeech/FenuMMM21,DBLP:conf/iccsa/FenuLM20,DBLP:conf/esse/FenuMMM20}. For instance, individuals with darker skin tones were disadvantaged by face authentication systems~\cite{8638319,9186002}. While research addressing this beyond accuracy objective is emerging, its scope has been restricted to pure biometric authentication and identification tasks~\cite{kleinberg2016inherent,atzori2022more}, leaving out potential adverse ramifications from a ranking standpoint and relying exclusively on close-range image data~\cite{demogpairs, terhorst2021comprehensive, unravelling,atzori2022explaining}.
From an information retrieval perspective, research on bias has predominantly centered on people ranking systems that do not rely on face biometrics~\cite{biega2018equity,singh2018fairness,lahoti2019operationalizing,DBLP:journals/ipm/BorattoFMM23,DBLP:conf/aied/FenuGM22,DBLP:conf/sigir/MarrasBRF22}. Forensic face rankings stand as a notably underexplored domain, distinguished by their complexity arising from normative considerations, content features, and architectural nuances.

In this paper, we aim to bridge the gap between the biometrics and information retrieval communities by assessing whether deep face recognition models contribute to unequal exposure among demographic groups in forensic ranking systems. Our novel contribution lies in two distinct aspects. Firstly, we propose an evaluation framework encompassing six cutting-edge face recognition models and two public face data sets, one annotated with gender and ethnicity labels and the other with only gender labels. Secondly, by simulating video surveillance re-identification and identification tasks, we conducted an exploratory study to address key research questions concerned with (i) identifying demographic groups prone to unfair rankings and exposure as well as (ii) quantifying disparate visibility and exposure based on the face probe's demographics. Our findings highlight biases within forensic rankings, underscoring the need for mitigation strategies for this task.

\section{Related Work}

Our work builds upon the key literature on face information forensics, face recognition in video surveillance scenarios, and bias in deep learning models for face recognition.

\subsection{Face Information Forensics}
Forensic face ranking methods had the tendency to rely on hand-drawn sketches from potential eyewitnesses~\cite{facesketch}. These partial details are then used in conjunction with other sources, e.g., textual descriptions ~\cite{hybridquery}. 
The integration of text-to-image and image-to-text techniques can make this strategy more effective, through the generation of queries and filtered galleries~\cite{imagetext}. 
During the identification process, the main objective is often to check the presence of the offender in the topmost position of the ranking produced by the method~\cite{multicrit}, disregarding the remaining positions. However, we believe that it is crucial to examine the ranking composition from a broader perspective and the (disparate) exposure of demographic groups since, for instance, being exposed as possible suspects in a ranking could have unintended implications (e.g., being erroneously investigated).

\subsection{Deep Face Recognition in Video Surveillance}
Face recognition within the context of video surveillance has predominantly prioritized efficiency and effectiveness. On one hand, the challenges behind low resolution and other variables encountered in uncontrolled scenarios complicate achieving optimal effectiveness. On the other hand, the demand for real-time monitoring mandates the prioritization of recognition speed. Modern solutions have addressed these objectives mainly through deep learning-driven knowledge distillation methodologies. For instance, \cite{knowledgedist1} established its foundation on a high-resolution teacher network to impart valuable knowledge for the training of a low-resolution student network. This approach has been refined by \cite{few} to encompass scenarios involving limited training instances. Conversely, \cite{Li2019OnLF} and \cite{yoon} adopted super-resolution techniques to reconstruct low-resolution images, a strategic choice driven by the substantial performance disparity between low- and high-resolution images. Additionally, strategies to address issues concerning pose and normalization have been explored in \cite{poseaware}, \cite{posenorm}, and \cite{poseinv}.

\subsection{Bias in Deep Face Recognition}
Previous research, as evidenced by~\cite{biasgenderbio,genderineq,sexist}, has demonstrated that gender biases in face recognition yield lower performance outcomes for women compared to men, on average. This gender-based discrepancy is frequently linked to the higher facial similarity among female faces, as highlighted in studies such as~\cite{DBLP:conf/interspeech/MarrasKMF19,genderineq,sexist}, as opposed to male faces. Additionally, age emerges as another demographic dimension fraught with systemic discrimination, manifesting in higher error rates for children's faces in contrast to those of adults. The potential for bias arising from an imbalanced representation of specific demographic groups has been acknowledged by~\cite{8638319,9186002}, prompting the development of demographically balanced data sets~\cite{inbook,9512390,9025435,demogpairs}. However, within these studies, there is notably no emphasis on the task of face recognition at a distance.

Despite concerted efforts to reduce data imbalances, face recognition models persist in their susceptibility to biases, and in certain instances, tend to internalize and amplify such biases~\cite{balancednotenough}. This prevailing concern has spurred researchers to delve into the origins of bias and utilize their findings to construct effective mitigation strategies. Substantial attention has been directed towards image-related factors~\cite{facequality}, encompassing distortions, noise, and facial attributes like makeup or mustaches. For example, the inadequate performance when processing images of individuals with darker skin tones or under low-light conditions can be attributed to the network's incorporation of skin-tone-related features in its upper layers. In a recent study by \cite{atzoriJSTSP}, biases within high- and low-resolution face recognition were examined through performance analysis in verification and identification. However, their investigation under the latter task did not consider the ranking composition. Despite these advances, a notable gap still affects the exploration of demographic disparities in the area of low-resolution, long-range face rankings, especially within video surveillance.

\begin{table*}[t!]
\label{table:3}
\begin{center}
\resizebox{\textwidth}{!}{%
\begin{tabular}{ll|rrr|rr|rr}
\hline
\textbf{Data Set Name}  & \textbf{Part}  & \textbf{\# Images} & \textbf{\# Users} & \textbf{Avg. Images/User} & \textbf{Avg. Images/Group} & \textbf{\# Groups} & \textbf{\% Minority}                   & \textbf{\% Majority}                       \\
\hline
DiveFace     & Training       & 111,503  & 19,196 & 6 $\pm$ 8           & 18,584 $\pm$ 2,968       & 6  & 49.07 (Male); 29.23 (Black)   & 50.92 (Female); 39.96 (Caucasian) \\
DiveFace     & Test       & 28,236  & 4,804 & 6 $\pm$ 11           & 4,706 $\pm$ 875       & 6  & 48.95 (Female); 27.72 (Black)   & 51.05 (Male); 39.91 (Caucasian) \\
VGG-Face2$^1$  & Test & 134,766 & 449  & 300 $\pm$ 108       & 22,461 $\pm$ 26,694         & 2  & 41.95 (Female) & 58.05 (Male) \\ 
\hline
\multicolumn{9}{l}{$^1$ Our experiments did not include models trained on \texttt{VGG-Face2}. For training, we used only \texttt{DiveFace}, which ensures a good representation of all demographic groups.} \tabularnewline
\end{tabular}
}
\end{center}
\vspace{-4mm}
\caption{Descriptive information about the two data sets considered in our study.}
\vspace{-5mm}
\label{tab:data}
\end{table*}     

\section{Problem Formulation}
In this section, we provide the necessary background to our methodology, by mathematically formulating the addressed tasks and defining the fairness notions of disparate visibility and exposure in the context of face rankings.

\vspace{2mm} \noindent \textbf{Face Recognition (Verification)}. Let $F \subset \mathbb{R}^{(*, *, 3)}$ denote the domain of three-channel color images. 
We consider a face encoder, parameterized by $\theta$, that yields face latent representations (embeddings) in $D_\theta \subset \mathbb{R}^e$,  where $e \in \mathbb N$. 
Formally, this step is denoted as $\mathcal{D}_{\theta}: F \rightarrow D_\theta$. 
Given a \textit{decision threshold} $\tau$, a verification trial can be defined as:

\vspace{-2mm}
\begin{equation} \label{eq:verification}
v_{\tau}: D_\theta \times D_\theta \rightarrow \{0,1\}    
\end{equation}

where a feature vector $d_p \in D_\theta$ from user $p$ is compared with a feature vector $d_u \in D_\theta$ from user $u$ to confirm (1) or confute (0) the assumption that the identities are the same. 

Given two fixed-length representations from a pair of images $(p,u)$, formally $d_p \in D_\theta$ and $d_u \in D_\theta$, and a \textit{decision threshold} $\tau \in \mathbb [0, 1]$, we describe the decision function implemented for a verification trial as: 

\vspace{-2mm}
\begin{equation} \label{eq:sim}
v_{\tau}(d_p , d_u) =  \cos(d_p, d_u) > \tau
\end{equation}   

where $cos$ is a Cosine similarity function, defined as:

\vspace{-5mm}
\begin{equation}
\resizebox{0.9\hsize}{!}{%
$\cos ({\bf d_p},{\bf d_u})= {{\bf d_p} {\bf d_u} \over \|{\bf d_p}\| \|{\bf d_u}\|} = \frac{ \sum_{i=1}^{n}{{\bf d_p}_i{\bf d_u}_i} }{ \sqrt{\sum_{i=1}^{n}{{\bf d_p}_i^2}} \sqrt{\sum_{i=1}^{n}{({\bf d_u}_i)^2}} }$
}
\end{equation}

Finally, obtaining a face verification model can be considered as an optimization problem aimed at maximizing the following expectation:

\vspace{-2mm}
\begin{equation} \label{eq:ex-func}
\underset{(\theta, \tau)}{\operatorname{argmax}} \mathop{\mathbb{E}}_{u, p \in F} 
\begin{cases}
    v_{\tau} \left( d_p, d_u \right) & \mathcal{I}(p) = \mathcal{I}(u) \\
    1 - v_{\tau} \left( d_p, d_u \right) & \mathcal{I}(p) \neq \mathcal{I}(u)
  \end{cases} 
\end{equation}

where $\mathcal{I}$ is a utility function that yields the identity depicted in the image passed as a parameter.

\vspace{2mm} \noindent \textbf{Face Recognition (Identification)}.
Given a set of individuals $\mathbb{I}$, let $P = \{p_1, ..., p_n\}$ be a set of images assumed to be probes during the identification task and $\bf{D} = \{d_1, ..., d_m \}$ be a set of the averaged embeddings saved in the gallery, one per each identity in $\mathbb{I}$, where $n = \lvert P \rvert$ and $m = \lvert \bf{D} \rvert$. Let ${R_i}$ be a list of identities in $\mathbb{I}$, sorted in decreasing order based on the Cosine similarity between the probe $p_i \in P$ and each averaged embedding $d \in \bf{D}$ in the gallery. For convenience, we denote the set of all rankings as $R = \{R_i \, | \, 1 \leq i \leq n \}$. Under these assumptions, obtaining a face identification model, parametrized by $\theta$, can be considered an optimization problem aimed to maximize the following expectation:

\vspace{-2mm}
\begin{equation} \label{eq:fnir}
\underset{\theta}{\operatorname{argmax}} \mathop{\mathbb{E}}_{1 \leq i \leq n} 
 v_{\tau} \left( d_{p_i}, d_{R_i[1]} \right) \cdot \mathbbm{1}(\mathcal{I}(p_i) = \mathcal{I}(R_i[1]))
\end{equation}

In other words, we expect that the Cosine similarity between the probe embedding $p_i$ and the topmost averaged embedding $R_i[1]$ is higher than the recognition threshold $\tau$ and that both images represent the same identity. 

\vspace{2mm} \noindent \textbf{Fairness Notions (Disparate Visibility and Exposure).} Following prior work in the information retrieval area~\cite{biega2018equity,singh2018fairness,lahoti2019operationalizing}, we consider disparate visibility and disparate exposure across demographic groups in the rankings as our fairness notions. The former is used to assess the likelihood of identities from a specific demographic group to appear at the top of the ranking, while the latter takes into consideration the specific \emph{position} at which identities from a demographic group are more likely to appear. 

Formally, let $A=\{a_1, a_2, ..., a_k\}$ be the set of demographic classes belonging to a given protected attribute $A$ each identity in the gallery is labelled with. We denote as $g \subset \mathbb{I}$ the set of identities that belong to a certain demographic group $a \in A$. In this way, we can define the visibility and exposure of a group $g$ in the rankings $R$ until a certain position $k$ as follows:

\begin{equation}
    \mathcal{V}(g, k) = \frac{1}{n \cdot k} \sum_{i=1}^{n} \sum_{pos=1}^{k} \mathbbm{1}(\mathcal{I}(R_i[pos]) \in g)
\label{eq:visibility}
\end{equation}

\begin{equation}
    \mathcal{E}(g, k) = \frac{1}{n \cdot o} \sum_{i=1}^{n} \sum_{pos=1}^{k} \frac{\mathbbm{1}(\mathcal{I}(R_i[pos]) \in g)}{log_2(pos+1)} 
\label{eq:exposure}
\end{equation}

where $o = \sum_{pos=1}^{k} 1 / log_2(pos+1)$ is the total decay associated with a ranking of length $k$; this term constrains the exposure estimates to lay in the range $[0, 1]$. Then, the disparate visibility (exposure) between two groups $g_i$ and $g_j$ until a certain position $k$ of the ranking can be defined as the absolute difference between the visibility (exposure) of the two groups. Formally, they can be defined as follows: 

\begin{equation}
\Delta \mathcal{V}(g_i, g_j, k) = | \mathcal{V}(g_i, k) - \mathcal{V}(g_j, k) |
\end{equation}

\begin{equation}
\Delta \mathcal{E}(g_i, g_j, k) = | \mathcal{E}(g_i, k) - \mathcal{E}(g_j, k) |
\end{equation}

Both disparity measures range between $0$ and $1$, with lower values meaning higher fairness (i.e., lower disparity). Note that the overall disparate visibility (exposure) of a model can be computed as the average disparate visibility (exposure) across all possible pairs of demographic groups.  

\section{Methodology}
Our methodology is based on four steps. First, we downloaded and pre-processed two public face data sets annotated with demographic labels (Tab. \ref{tab:data}). Second, we degraded each image to simulate its counterpart collected at a distance. Third, we trained well-known face encoders with either the original or the degraded counterparts of the images belonging to the identities in the training set. Finally, we assessed the effectiveness and fairness of these face encoders, according to the formulation provided above.

\vspace{2mm} \noindent \textbf{Data Preparation}. We conducted our experiments using the DiveFace \cite{diveface} and VGGFace2 \cite{cao2018vggface2} data sets. The former includes 140,000 images associated with 24,000 identities.  It is appropriately annotated with sensitive data (gender and ethnicity) and balanced in terms of both characteristics. 
Ethnicity labels include Asian, Black, and Caucasian. Gender labels include Women and Men. Therefore, the data set contains six demographic groups: Asian Men, Asian Women, Black Men, Black Women, Caucasian Men, and Caucasian Women.
The original authors divided the entire data set into two sets, a training set and a test set, containing 70\% and 30\% of the identities, respectively.
This data set is, to the best of our knowledge, the state-of-the-art (non synthetic) source for fairness analysis in face biometrics.
The latter contains 3.3 million images representing over 9,000 identities. 
Due to its size, only the original test set of 134,766 images from 449 identities was included in our analyses.
Later, demographic annotations were added to \cite{maad1, maad2} (MAAD-Face). 
This data set contains only binary gender labels for Women and Men. DeepFace~\cite{serengil2020lightface} was used 
to detect, crop, and resize faces in the original images.

\vspace{2mm} \noindent \textbf{Face Image Degradation}. Using unpaired long- and close-range images, following the protocol proposed by \cite{atzoriJSTSP}, we trained a near-to-far GAN that efficiently learns how to carry out the image degradation process (unpaired image-to-image translation job) and, therefore, emulate face pictures captured by surveillance cameras at a distance. To train our near-to-far network, we used QMUL-SurvFace \cite{qmul} as long-range data set, since its images have been acquired by real-world surveillance cameras at various distances. The close-range data set, as in \cite{yoon}, was obtained by combining the whole AFLW data set \cite{AFLW} (more than 20,000 faces in various poses and expressions), a subset of LS3D-W \cite{ls3d} (faces with large variation in terms of pose, illumination, expression and occlusion), and a subset of the VGGFace2 training set. To ensure that the degradations carried out by the near-to-far GAN would only acquire artifacts, turbulence, and noise in the reference long-range images (otherwise, the restoration step would not have produced a congruent image), we simultaneously trained a far-to-near GAN to restore the original image from the artificially degraded one. Its discriminator has the task of differentiating the restored image from the original close-range image given as input to the first GAN. The quality of the degradation process has been assessed in \cite{atzoriJSTSP}. 

\vspace{2mm} \noindent \textbf{Face Encoders Preparation}. In this step, we reproduced six state-of-the-art convolutional neural networks (CNNs), namely \texttt{MobileFaceNet}~\cite{mobilefacenet}, 
\texttt{ResNet152}~\cite{he2015deep}, 
\texttt{AttentionNet}~\cite{wang2017residual}, 
\texttt{ResNeSt}~\cite{zhang2020resnest}, 
\texttt{RepVGG}~\cite{ding2021repvgg}, and
\texttt{HRNet}~\cite{wang2020deep}). 
Recent benchmarks proved that these architectures can achieve state-of-the-art performance ~\cite{wang2021facex}.
Each CNN has been extended with a classification head that computes a Negative-Positive Collaboration (NPCFace) loss function \cite{zeng2020npcface}, which incorporates the improved hard negative emphasis and the collaboration in hard positive emphasis.
Subsequently, we trained six deep face encoders, one for each CNN architecture, using face images from the DiveFace \cite{diveface} training set, degraded or not based on the considered scenario. 
For the sake of consistency, as in \cite{atzoriJSTSP}, each encoder was trained with 64-sized batches for a maximum of 80 epochs (early stopping, patience 5). The optimizer was SGD with momentum of 0.9, weight decay of 1e-8, an initial learning rate of 0.10 and decays at 5, 25, and 68 epochs, under a categorical cross-entropy.

\begin{figure}[t]
\begin{subfigure}{\linewidth}
\centering
    \includegraphics[width=.9\linewidth, valign=t]{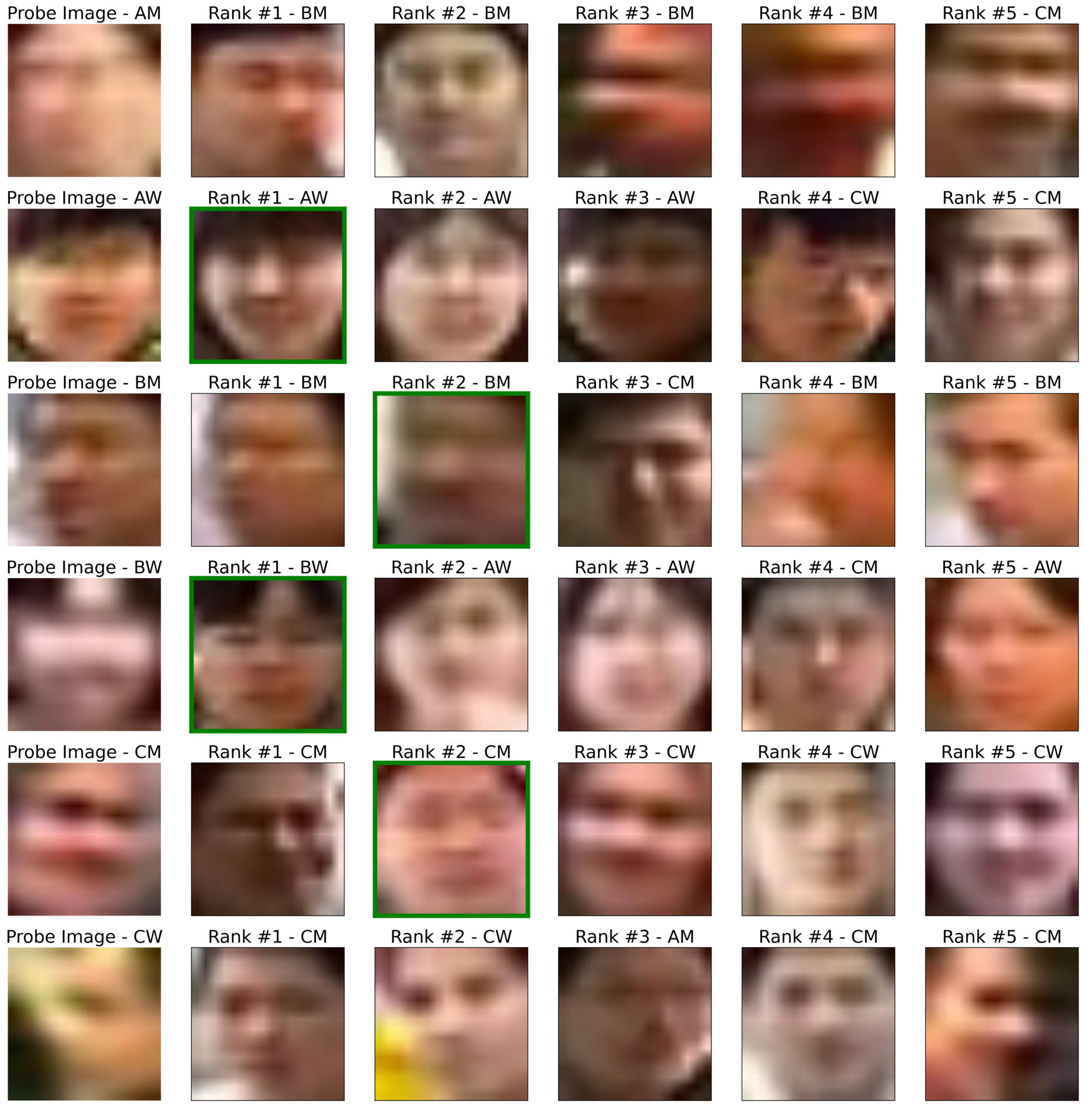}
    \caption{Rankings under long-range galleries.}
    \label{fig:reid}
\end{subfigure}
\begin{subfigure}{\linewidth}
\centering
    \includegraphics[width=.9\linewidth, valign=t]{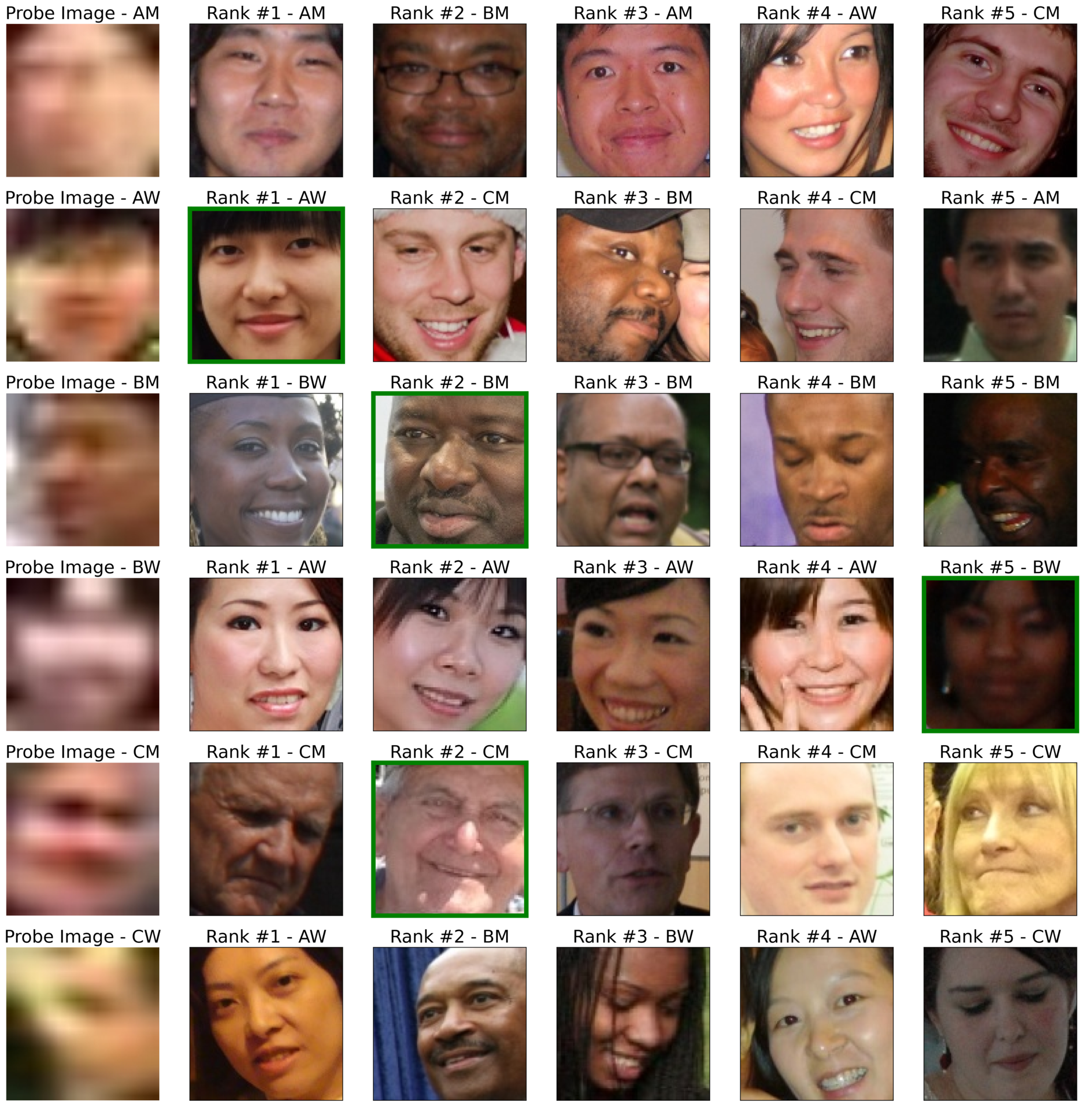}
    \caption{Rankings under close-range galleries.}
    \label{fig:id}
\end{subfigure}
\vspace{-5mm}
\caption{Example rankings for both long- and close-range galleries. Each image reports its ranking position and demographic group. Green-border images are probe's real mates. \label{fig:rankex}}
\vspace{-6mm}
\end{figure}

\begin{figure*}[!t]

\begin{subfigure}{\linewidth}
\centering
    \includegraphics[width=\linewidth, valign=t]{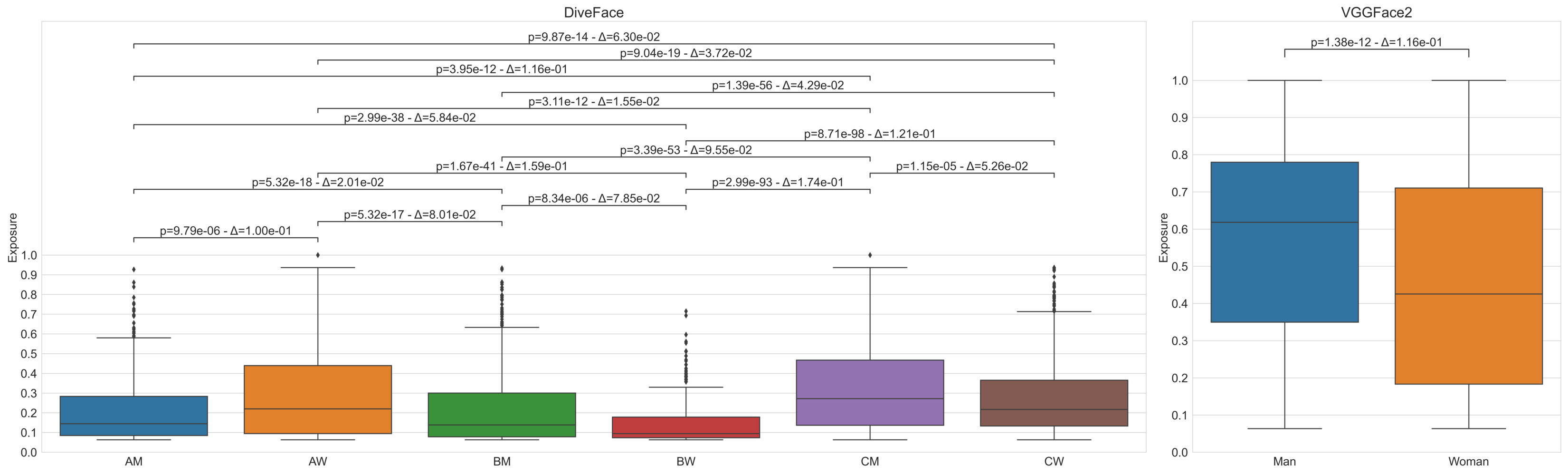}
    \caption{Rankings under long-range galleries.}
\end{subfigure}
\begin{subfigure}{\linewidth}
\centering
    \includegraphics[width=\linewidth, valign=t]{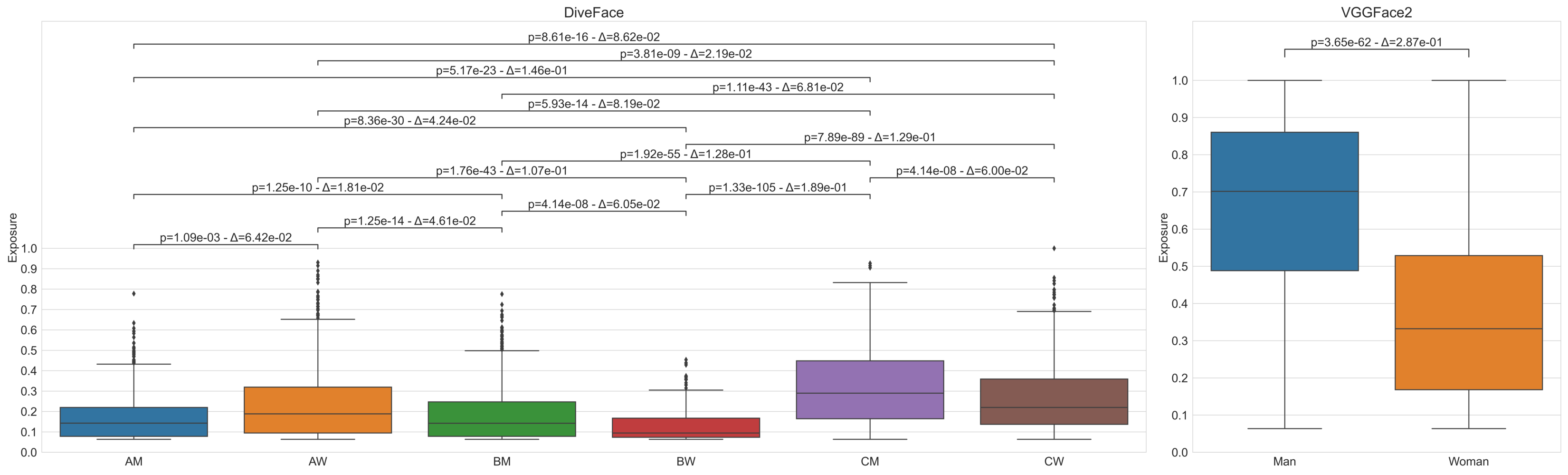}
    \caption{Rankings under close-range galleries.}
\end{subfigure}
\vspace{-7mm}
\caption{Groups exposure averaged across face encoders. Statistical significance assessed via a Kolmogorov-Smirnov test, which rejects the hypothesis that exposure distributions are identical between the groups, if the p-value is lower than $0.05$. \label{fig:overallEXP}}
\end{figure*}

\vspace{2mm} \noindent \textbf{Rankings Generation}.
With the face images from each test set, we developed a ranking system that, given the latent representation of a long-range image of an individual (probe) as a query, ranks all the identities in the gallery and returns the $k=10$ identities most similar to the query. For this purpose,  we only considered individuals with at least $l=10$ face images in the test sets and sampled $l$ images for each individual. Given $l$ images for an individual, $30\%$ were assumed to be face probes (images to be used as a query), while $70\%$ made up the gallery. Each gallery was produced by averaging every latent representation in this set, in order to achieve a single, invariant latent representation of the individual. For each probe, we computed the Cosine similarity between its latent representation and the identities' ones. We ranked the identities in the gallery by their decreasing similarity with the probe and only considered the $k$ identities most similar to the query, for each ranking. Long-range galleries, as exemplified in Fig. \ref{fig:reid}, were used to simulate a re-identification task (both the probe and the gallery at a distance), while close-range ones (Fig. \ref{fig:id}) were considered for an identification task (only the probe at a distance).

\begin{figure*}[t]%
    \centering
    \subfloat[DiveFace]{
    \includegraphics[width=.48\linewidth, trim=4 4 4 4,clip]{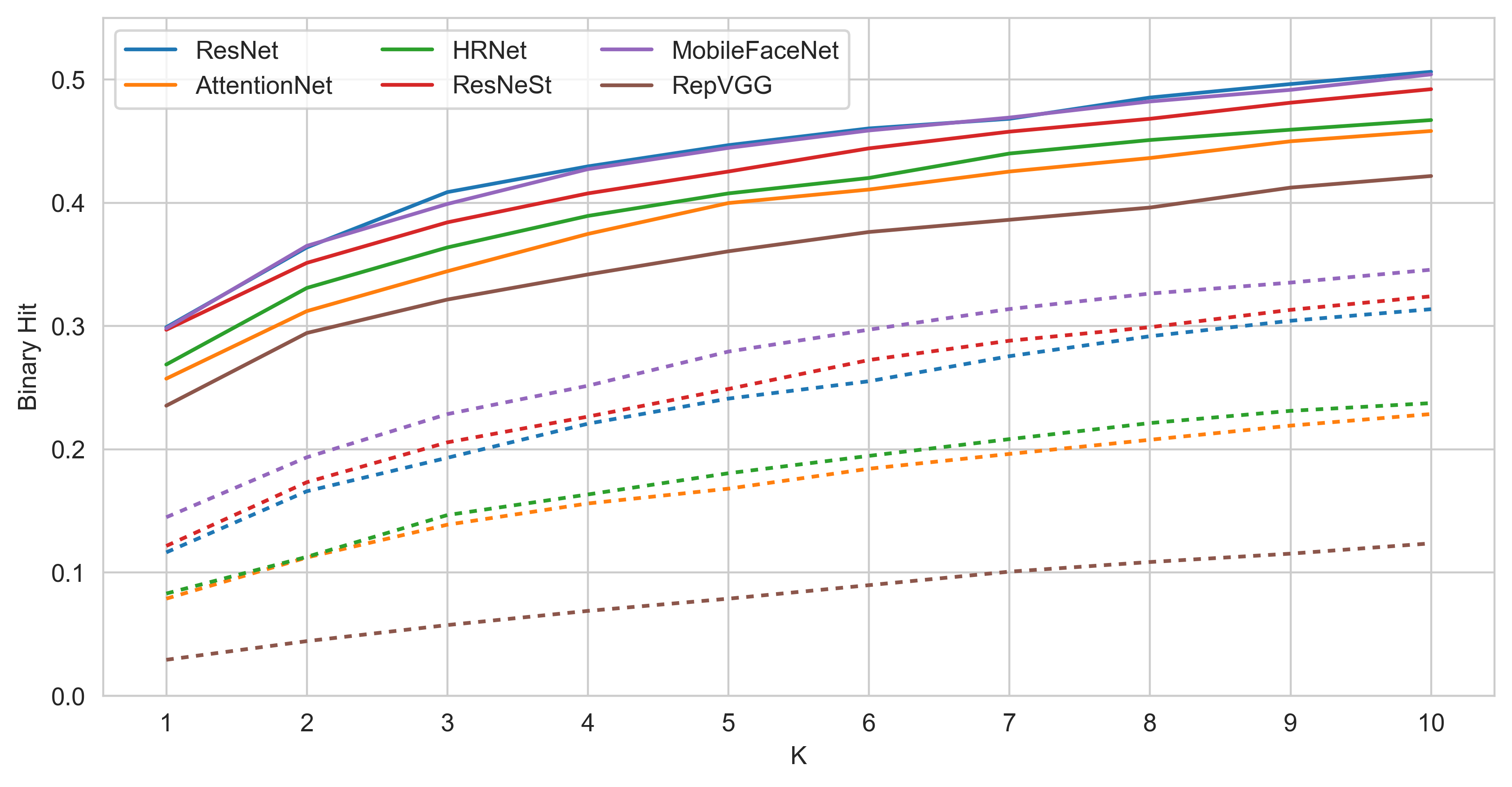}}
    \subfloat[VGGFace2]{
    \includegraphics[width=.48\linewidth, trim=4 4 4 4,clip]{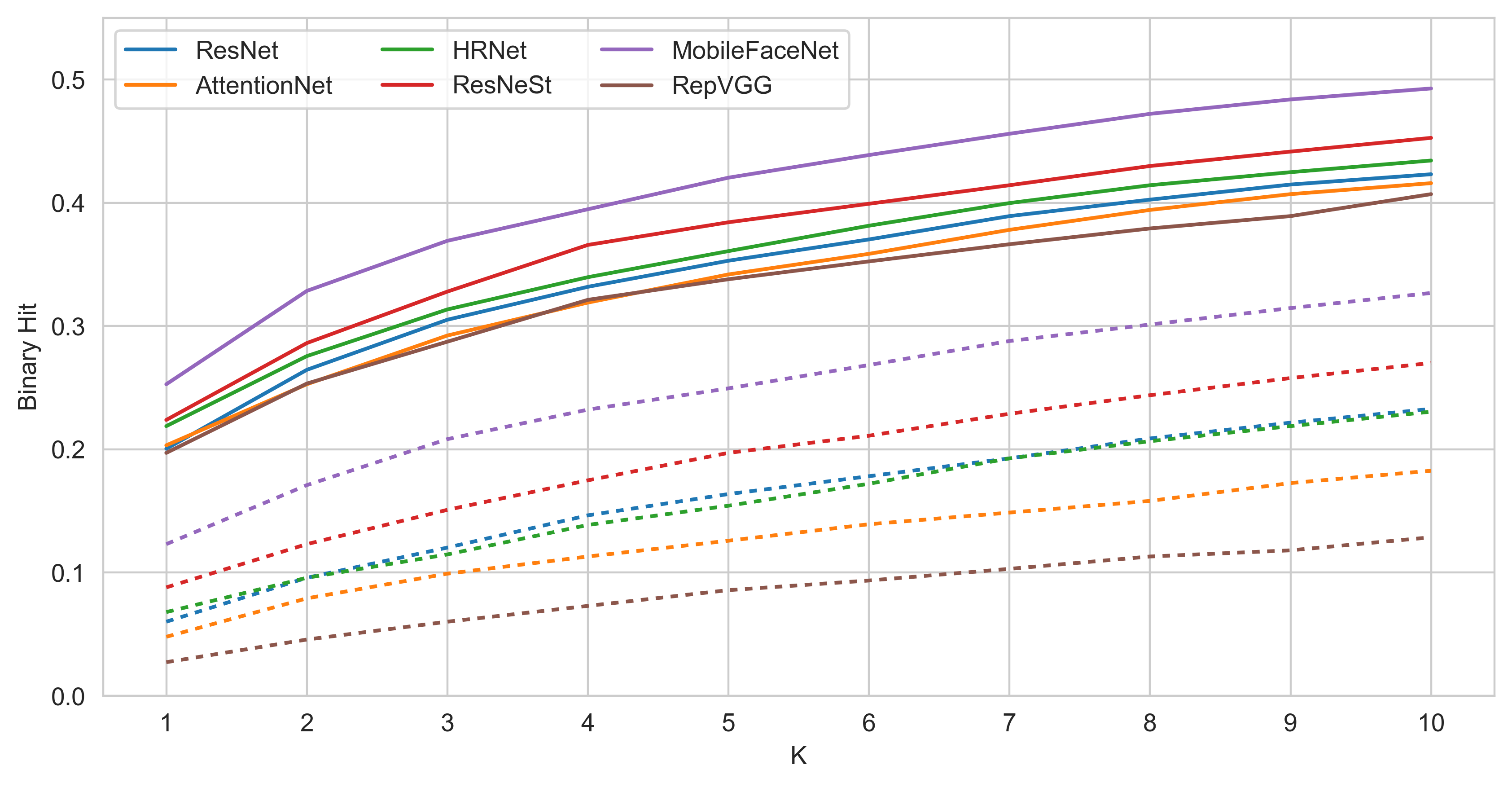}}
    \vspace{-3mm}
    \caption{Model performance in terms of binary hit ratio, namely the percentage of rankings on which the probe's identity is present in at least one of the first $k$ positions. Solid lines refers to long-range galleries, dashed lines to close-range galleries. \label{fig:modelperf}}
\end{figure*}

\section{Experimental Results}
Our experiments analyzed demographic group's overall exposure in rankings (RQ1), visibility per probe's group (RQ2), and exposure according to the probe's group (RQ3). 

\subsection{Overall Disparate Exposure (RQ1)}
\label{sec:rq1}

In a first analysis, we investigated whether the exposure across demographic groups is evidently disparate if we consider the position in which a certain possible suspect might appear. Since our results were shown to be consistent across models, Fig. \ref{fig:overallEXP} collects exposure distributions of groups over rankings, averaged across face encoders. 

Comparing exposure across groups, it can be observed that certain groups tend to have a statistically significant disparate exposure compared to other groups. Our results were strongly consistent with both close- and long-range galleries. In decreasing order, on the \texttt{DiveFace} data set, the most exposed demographic groups on long- (close-) range galleries were Caucasian Males, Asian Women (Caucasian Women), Caucasian Women (Asian Women), Asian Males, Black Males, and Black Women. Notably, except for Asians, Males were reporting higher exposure values than Women from the same ethnicity. In both settings, Black Women reported the lowest exposure values, indicating how their false positives are less likely to take prominent positions in the rankings. On the other side, Caucasian Men appear to be the most likely demographic group to appear in higher positions of the rankings. Indeed, on the \texttt{VGGFace2} data set, Males got higher exposure under both gallery ranges (but more notably in the close-range one).

\hlbox{RQ1 Summary}{Except for Asians, Women tend to have a lower exposure than Males. Black Women got the lowest exposure across all groups, on average.} 

\subsection{Disparate Visibility per Probe’s Group (RQ2)}
\label{sec:rq2}

In a second analysis, for each probe's group, we computed the visibility of each demographic group across rankings. In this scenario, our experiments were focused on \texttt{MobileFaceNet}, since it achieved, as shown in Fig. \ref{fig:modelperf}, the best results under both long- and close- range galleries.

\begin{figure*}[!t]%
    \centering
    \hspace{0.8cm}
    \subfloat[Long-range galleries]{
    \includegraphics[width=.5\linewidth, trim=4 4 4 4,clip]{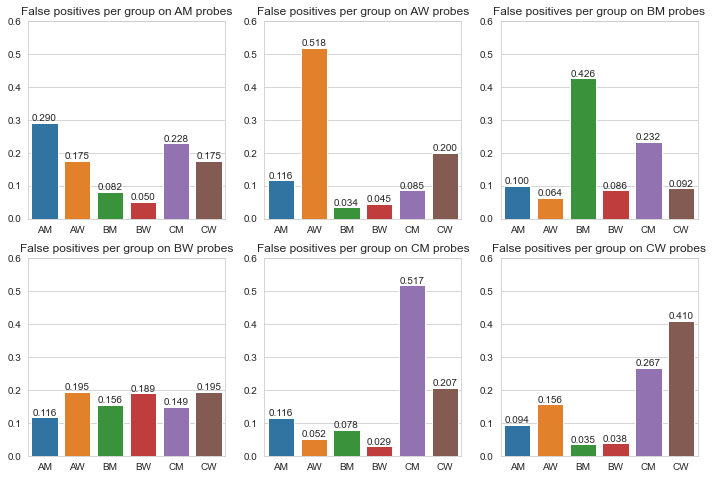}}
    \subfloat[Close-range galleries]{
    \includegraphics[width=.5\linewidth, trim=4 4 4 4,clip]{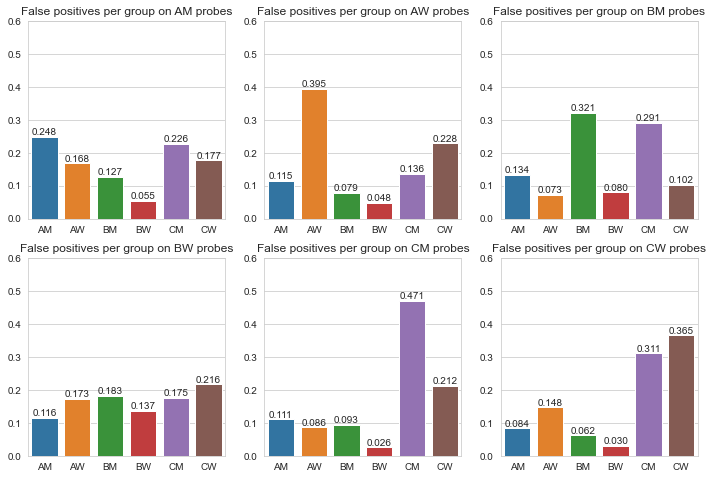}}
    \caption{Demographic groups visibility (i.e., false positive ratio) in top-10 rankings for each probe’s group. For the sake of conciseness, we adopted the legend A:Asian; B:Black; C:Caucasian; M:Male; W:Women. \label{fig:falsepositives}}
\end{figure*}

Observing Fig. \ref{fig:falsepositives}, as expected, for each probe's group, the demographic group with the highest visibility corresponds to the probe's one (visibility level between 29\% and 52\%).
 Going further, probes of Asian men (top left quadrants) led to higher disparate visibility for Caucasian men (same gender) and Asian women (same ethnicity) than the other groups. Under both close- and long-range scenarios, Caucasian Women surprisingly obtained high visibility: in the long-range one, they got the exact same score as Asian Women (same ethnicity), while in the close-range one, they got a higher visibility (+0.9\%). Probes of Asian Women led to a similar result, with a disparate visibility for Caucasian Women (same gender) and Asian Men (same ethnicity). Similarly to their Women counterpart, under close-range galleries, these probes led also to high visibility for Caucasian Men (+2.1\% w.r.t. Asian Men). 
 
 Black Men probes led, in both close- and long-range galleries, to disparate visibility for same-gender groups, namely Caucasian Men and Asian Men, leaving their female counterpart in the second to last position. Black Women probes led to the fairest result among our experiments, showing a strong predisposition to be matched by same-gender groups (Caucasian Women and Asian Women, in order). In these scenarios, both groups obtained an higher visibility than the probes' group. Also, Black Women probes led to the lowest same-group visibility scores, showing better suitability for ranking approaches. Caucasian Men probes showed the exact same behaviour in both types of gallery ranges, leading to disparate visibility for Caucasian Women (same ethnicity). Lastly, Caucasian Women probes led to disparate visibility for Caucasian Men (same ethnicity) and Asian Women (same gender).

\hlbox{RQ2 Summary}{Depending from the probe’s group, certain groups (beyond that of the probe) are disproportionately visible at the top of the rankings, particularly Caucasian Male and Asian Women.} 

\begin{figure*}[t!]%
    \centering
    \hspace{0.8cm}
    \subfloat[Long-range galleries]{
    \includegraphics[width=.5\linewidth, trim=4 4 4 4,clip]{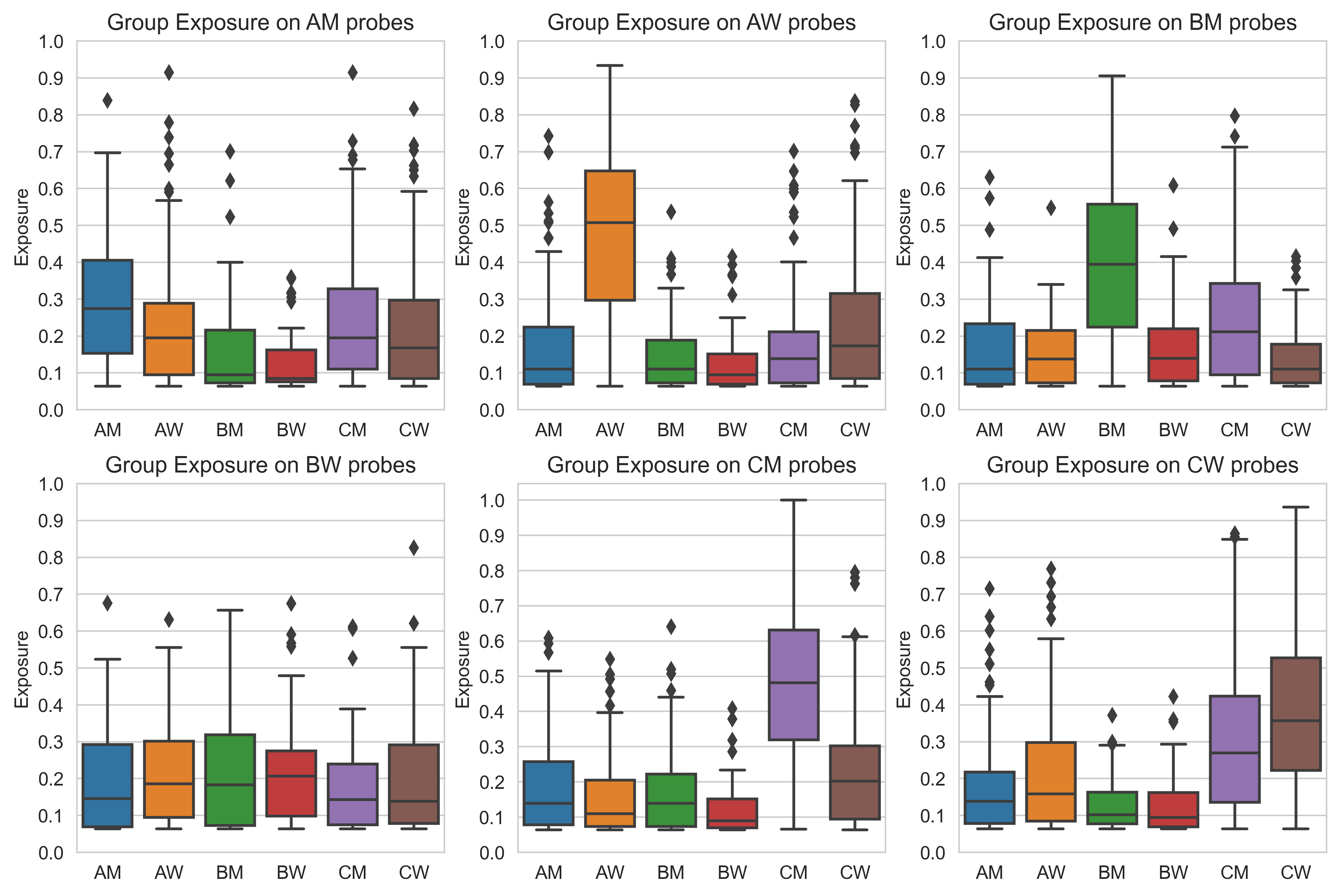}}
    \subfloat[Close-range galleries]{
    \includegraphics[width=.5\linewidth, trim=4 4 4 4,clip]{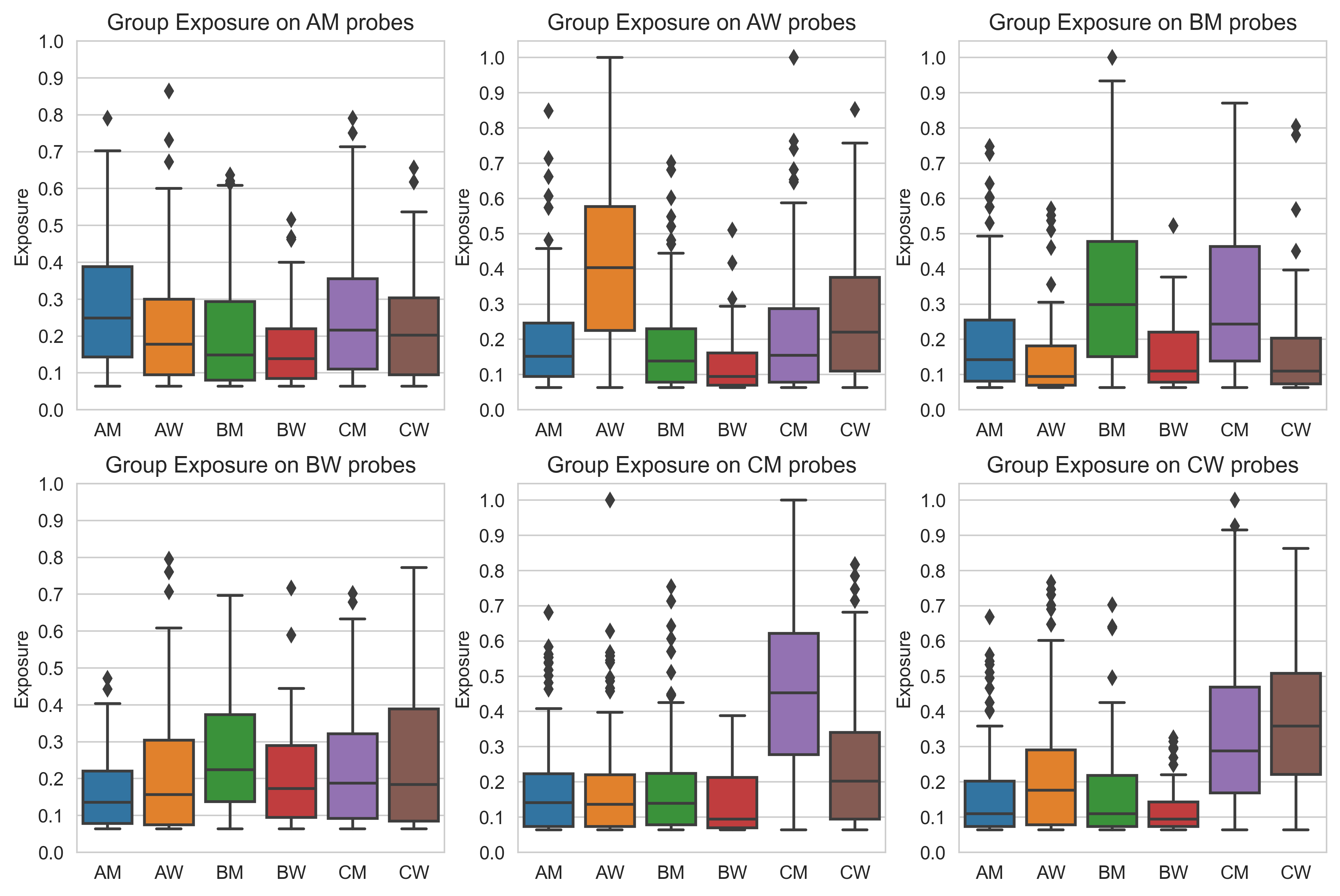}}
    \caption{Demographic groups exposure in top-10 rankings for each probe’s group. For the sake of conciseness, we adopted the legend A:Asian; B:Black; C:Caucasian; M:Male; W:Women. \label{fig:groupexp}}
\end{figure*}

\subsection{Disparate Exposure per Probe’s Group (RQ3)}
In our third analysis, we investigated whether the average exposure of groups changes based on the probe's group. Indeed, the analysis on the averaged exposure performed in Section \ref{sec:rq1} does not allow to make any conclusion on the susceptibility of exposure from the probe's group. Furthermore, we were interested in understanding whether the trends observed for the visibility estimates according to the probe's group in Section \ref{sec:rq2} are confirmed also in case we consider group exposure as fairness notion. 

Fig. \ref{fig:groupexp} shows that the face encoders are susceptible to rank identities from the same group of the probe in higher positions. For instance, Asian Men probes tend to yield higher exposure with Caucasian Men (same gender), Caucasian Women (no common attributes and only with long-range probes), and Asian Women (same ethnicity). Asian Women, instead, strictly follow the previous observed visibility distribution, obtaining higher exposure for Caucasian Women (same gender), Asian Men (same ethnicity), and Caucasian Men (no common attributes). With close-range galleries, however, their Male counterpart tend to have lower exposure than the other mentioned groups. Black Males probes led to disparate exposure for Caucasian Men and Asian Men (both with the same gender). Black Women probes led to the lowest disparity estimates among our results (as in Section \ref{sec:rq2}), with higher exposure for their Male counterparts. In long- (close-) range settings, the group with the lowest exposure for these probes were Caucasian (Asian) Men, while same-group galleries were always less exposed than all the other groups. Caucasian Men probes led (only close-range) to disparate exposure for their Women counterparts. The same behavior was observed for Caucasian Women (for their Male counterpart), while they led to disparate exposure for Asian Women (same gender).

\hlbox{RQ3 Summary}{Group visibility trends were confirmed for group exposure as well: highly visible groups are also highly exposed, i.e., they surpass other groups both in quantity and ranking placement.} 


\section{Conclusions and Future Work}
In this paper, we investigated the extent to which state-of-the-art face encoders adopted for forensic face ranking are subject to biases across demographic groups. Due to the lack of long-range data sets annotated with demographic attributes, our protocol employed near-to-far GANs to instill noise and degradation due to distance and atmospheric conditions into the close-range, originally annotated images.

Our results highlighted that Caucasians Males and Asian Women are disproportionately exposed in the rankings, especially under close-range settings. Differently from prior work, dark-skinned individuals (particularly Women) are less exposed under probes belonging to other groups. We conjecture that dark-skinned groups, by means of a face ranking system, tend to follow a more uniform distribution of false positives across probe groups. This points out that especially Black Women are less likely to be erroneously investigated, while Caucasians and Asians, showing higher intra- and inter-group similarities, are more likely to be.

In the next steps, we plan, using a wider range of state-of-the-art data sets, to investigate whether, and possibly to what extent, other factors (e.g., pose, noise, lighting, facial hair) influence forensic face rankings, addressing also the limitations concerning the adoption of GANs for simulating recognition at a distance. Finally, we also plan to devise countermeasures against the uncovered disparities.

\vspace{2mm} \noindent \textbf{Acknowledgements}. We acknowledge financial support under the National Recovery and Resilience Plan (NRRP), Mission 4 Component 2 Investment 1.5 - Call for tender No.3277 published on December 30, 2021 by the Italian Ministry of University and Research (MUR) funded by the European Union – NextGenerationEU. Project Code ECS0000038 – Project Title eINS Ecosystem of Innovation for Next Generation Sardinia – CUP F53C22000430001- Grant Assignment Decree No. 1056 adopted on June 23, 2022 by the Italian Ministry of University and Research.




\balance

{\small
\bibliographystyle{ieee}
\bibliography{egbib}
}

\end{document}